\documentclass{article}

\usepackage{arxiv}

\usepackage[utf8]{inputenc} % allow utf-8 input
\usepackage[T1]{fontenc}    % use 8-bit T1 fonts
\usepackage{url}            % simple URL typesetting
\usepackage{booktabs}       % professional-quality tables
\usepackage{amsfonts}       % blackboard math symbols
\usepackage{nicefrac}       % compact symbols for 1/2, etc.
\usepackage{microtype}      % microtypography
\usepackage{lipsum}		% Can be removed after putting your text content
\usepackage{graphicx}
\usepackage{natbib}
\usepackage{doi}
\usepackage{subfig}  % for using subfloats in pictures
\usepackage{multirow}
\usepackage{caption}
\usepackage{paralist, tabularx}
\usepackage[blocks]{authblk}

\newcommand{\tabitem}{~~\llap{\textbullet}~~}

\usepackage[final]{changes}

\title{Does a Technique for Building Multimodal Representation Matter? -- Comparative Analysis}

\date{} 					% Or removing it

\author[1]{\href{https://orcid.org/0000-0002-3096-9918}{Maciej Pawłowski}}
\author[1]{\href{https://orcid.org/0000-0002-3407-7570}{Anna Wróblewska}}
\author[2]{\href{https://orcid.org/
0000-0001-5960-8131}{Sylwia Sysko-Romańczuk}}

\affil[1]{Faculty of Mathematics and Information Science, Warsaw University of Technology, Warsaw, Poland \linebreak
\texttt{mac.pawlowski19@gmail.com, anna.wroblewska1@pw.edu.pl}}
\affil[2]{Faculty of Management, Warsaw University of Technology, Warsaw, Poland\\ 
\texttt{sylwia.sysko.romanczuk@pw.edu.pl}}

\renewcommand{\headeright}{}
\renewcommand{\undertitle}{}
\renewcommand{\shorttitle}{Does a Technique for Building Multimodal Representation Matter?}

\hypersetup{
pdftitle={Does a Technique for Building Multimodal Representation Matter? -- Comparative Analysis},
pdfsubject={cs.AI,cs.LG},
pdfauthor={Maciej Pawłowski, Anna Wróblewska, Sylwia Sysko-Romańczuk},
pdfkeywords={Multimodal Representation, Multimodal Learning, Data Fusion, Comparative Analysis},
}

\begin{document}
\maketitle
\begin{abstract}
    Creating a meaningful representation by fusing single modalities (e.g., text, images, or audio) is the core concept of multimodal learning. Although several techniques for building multimodal representations have been proven successful, they have not been compared yet. Therefore it has been ambiguous which technique can be expected to yield the best 
    results in a given scenario and what factors should be considered while choosing such a technique. This paper explores the most common techniques for building multimodal data representations -- the late fusion, the early fusion, and the sketch, and compares them in classification tasks.  Experiments are conducted on three datasets: Amazon Reviews, MovieLens25M, and MovieLens1M datasets. In general, our results confirm that multimodal representations are able to boost the performance of unimodal models from $0.919$ to $0.969$ of accuracy on Amazon Reviews and $0.907$ to $0.918$ of  AUC on MovieLens25M. However, experiments on both MovieLens datasets indicate the importance of the meaningful input data to the given task. In this article, we show that the choice of the technique for building multimodal representation is crucial to obtain the highest possible model's performance, that comes with the proper modalities combination. Such choice relies on: the influence that each modality has on the analyzed machine learning (ML) problem; the type of the ML task; the memory constraints while training and predicting phase.
\end{abstract}

% keywords can be removed
\keywords{Multimodal Representation, Multimodal Learning, Data Fusion, Comparative Analysis}

\section{Introduction}\label{sec:intro}

Multimodal learning involves working with data that contains multiple modalities such as text, images, audio, numerical or behavioral data. 
The interest in multimodal learning started in the 1980s and has gained popularity since then. In one of the first works on the subject~(\cite{yuhas_integration_1989}), the authors demonstrated that acoustic and visual data could be successfully combined in the speech recognition system.  They showed that integrating these modalities outperforms the audio data model.

Research shows that modalities are complementary and multimodal learning should be explored comprehensively, as it can significantly enhance models' performance in any machine learning task~(\cite{baltrusaitis_multimodal_2019,cao_review_2020,gao_survey_2020}).
So far, a multimodal approach has been proven effective in numerous fields, i.e., in information retrieval~(\cite{cao_review_2020}), medicine~(\cite{el-sappagh_multimodal_2020}) and human behavior analysis~(\cite{jaiswal_muse:_2020}). 

The central concept of multimodal learning is to develop a data fusion technique that can be applied to numerous machine learning tasks with a few chosen or all kinds of modalities (universal technique). According to~\cite{gao_survey_2020}, such a universal approach has not been established so far. In their work, the authors noticed that all current state-of-the-art data fusion models are suffering from some issues. Either they are task-specific or too complicated, in the sense that they lack interpretability and flexibility. 

We believe that in order to develop such a universal technique, it is necessary to understand the differences between the existing methods. To the best of our knowledge, we are the first to investigate the behaviors of multimodal data fusion techniques in several scenarios. In this paper, we show that the effective multimodal representation depends on the type of technique used to build this representation and other criteria/requirements. We distinguish three main criteria that define this dependency: the influence that each modality has on the analyzed machine
learning (ML) problem; the type of the ML task; the memory
constraints while training and predicting phase.

% contributions
This work contributes to examining the most common techniques of building multimodal data representations: the late fusion, the early fusion, and the sketch.
% \footnote{The tested dataset preprocessing, and source code will be opened after acceptance.}
In the first approach -- the late fusion -- all the modalities are learned independently and are combined right before the model makes a decision. The second technique -- the early fusion -- first combines all modalities and then the model is trained. The last one -- the sketch -- is similar to the early fusion; however, modalities are transformed into a common space instead of being concatenated. All of these techniques have been proven effective. In this paper, we perform this comparison in classification tasks. Finally, we define criteria and recommendations on how to choose the multimodal technique. The criteria comprise: the influence that each modality has on the analyzed machine learning (ML) problem, the ML task type, the memory constraints while the training and predicting phase.
%, the possibility to speed up training with transfer learning (using pre-trained models). 
 In our work we use three datasets -- Amazon Reviews~(\cite{he_ups_2016}), MovieLens25M and MovieLens1M~(\cite{harper_movielens_2016}). They encapsulate three modalities -- textual, visual, and graph data; each can bring unique information to the ML model. 

Thus, in the following, in Section~\ref{section:related_works} we state the problems of multimodality, and present employed datasets (Section~\ref{section:datasets}). Finally, we experiment with the chosen multimodal approaches (Section~\ref{section:research_design} and \ref{section:results}), and provide a few conclusions in Section~\ref{section:conclusions}.

\section{Related Work}
\label{section:related_works}

\subsection{Types of Modalities}

\begin{figure}[!ht]
    \centering
    \includegraphics[width=\columnwidth]{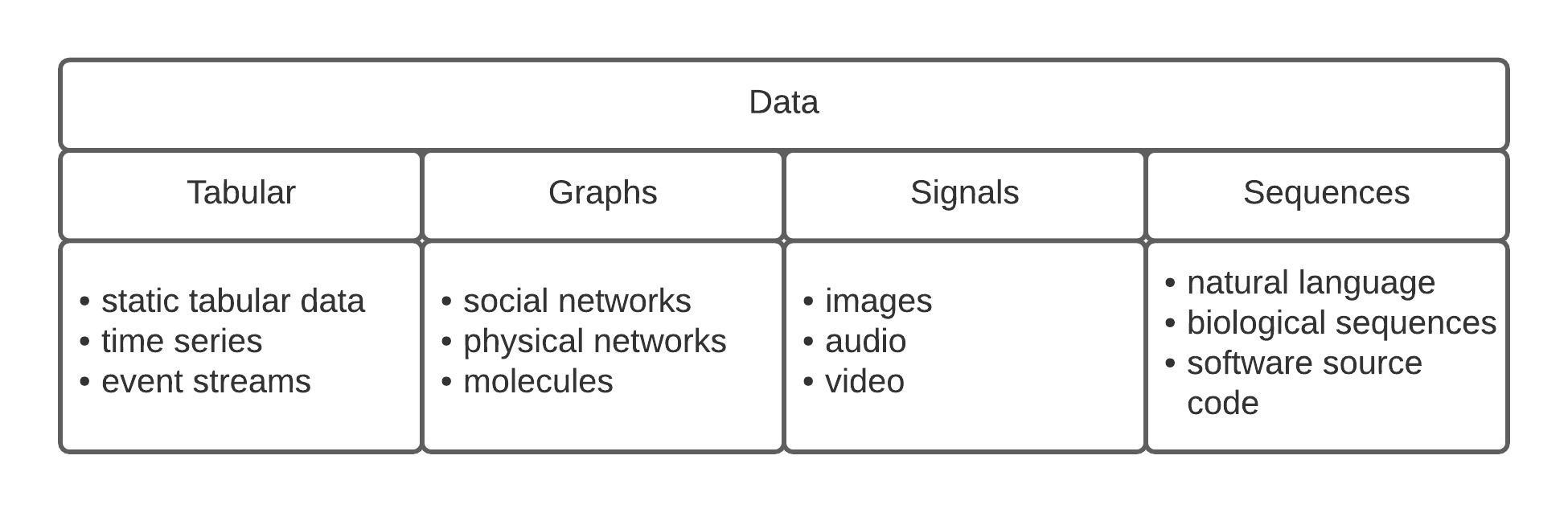}
    \caption[Types of modalities]%
    {Types of modalities that can be encountered. The division comes from~\cite{varshney_trustworthy_2022}.}
    \label{fig:modalities_types}
\end{figure}

This section introduces the types of modalities that can be encountered while working on any machine learning problem. Figure~\ref{fig:modalities_types} depicts the full spectrum of existing modalities. Accordingly, to~\cite{varshney_trustworthy_2022}, we identify 4 four main groups of modalities:
\begin{itemize}
    \item tabular data -- observations are stored as rows, and their features as columns;
    \item graphs -- observations are vertices, and their features are in the form of edges between individual vertices;  
    \item signals -- observations are files of appropriate extension (images - .jpeg, audio - .wav, etc.), their features are the numerical data provided within files;
    \item sequences -- observations are in the form of characters/words/documents, where the type of characters/words corresponds to features.
\end{itemize}

Our study includes modalities coming from each of the identified groups: labels are an example of tabular data, reviews, titles, descriptions represent textual data (sequences), images of products, and movie posters are examples of visual data, and relations between movies (movies seen by one user) are graphs. Table~\ref{tab:examples_modalities} introduces some other examples of works on various combinations of modalities. Additionally,~\cite{ZHANG2021104042} concentrate on the image segmentation task to learn the optimal joint representation from rich and complementary features of the same scene based on images retrieved from different sensors. All these studies show that there are no restrictions on modalities that can be studied together to solve a specific problem. Moreover, they prove that the combination of modalities boosts performance, that is, the multimodal model achieves better results than unimodal models. 

Furthermore, these works encapsulate all multimodal fusion techniques, that we examine in our research, the early fusion, the late fusion, and the sketch.  All of them are proven effective but have not been compared until our work.

\begin{table}[!ht]
\centering
\begin{tabularx}{\textwidth}{>{\raggedright\arraybackslash}p{2cm}p{3.7cm}X}
\toprule
Modalities & Article & Overview \\ \midrule
images; time series; tabular
& \cite{el-sappagh_multimodal_2020} & Prediction of Alzheimer's disease based on magnetic resonance imaging, positron emission tomography (images) that are performed multiple time on one patient within specified periods of time (time series). Patient demographics, and genetic data are also taken into account (tabular). \\
 \midrule
 audio; video; event streams & \cite{jaiswal_muse:_2020} & Behavioural analysis - emotion, and stress prediction. Analysed data consist of 45-minutes recordings of students during final exam period. They are recorded with the use of cameras (video), thermal physiological measurements of heart and breathing rates (event streams), lapel microphones (audio). \\ 
 \midrule
 text; images
 & \cite{singh_towards_2019} & Question answering based on images containing some textual data. \\ 
 \midrule
 images; text; graphs & \cite{rychalska_i_2020} \ \ \  \cite{LAENEN2020102316} \cite{JMLR:v21:19-805}& Outfit / Movie recommender systems. Movies are recommended based on plot (text), poster (image), liked and disliked movies, cast (graphs). Outfits are chosen based on product features in images and text descriptions. \\
\bottomrule
\end{tabularx}
\caption[Examples of multimodal tasks]{Examples of multimodal tasks}
\label{tab:examples_modalities}
\end{table}

\subsection{Multimodal Representation}
Learning to represent is an unsupervised task, and there is no single manner to describe what a good representation is. However, several works have already identified the main features demanded while deriving any numerical representation of a given modality.

The problem of unimodal representation has been already solved with modality-dedicated models, such as BERT~(\cite{devlin_bert:_2019}) for textual data, ResNet~(\cite{he_deep_2016}) for images, etc. However, there has not been established~(\cite{gao_survey_2020}) a universal method that could be applied to any machine learning task when it comes to multimodal data. 

\cite{bengio_representation_2013} characterize several features that an appropriate vector representation should possess, including:
\begin{itemize}
    \item smoothness -- transformation should preserve objects similarities, mathematically  $x \approx y \Rightarrow f(x) \approx f(y)$. For instance, the words "book" and "read" are expected to have similar  embeddings;
    \item manifolds -- probability mass is concentrated within regions of lower dimensionality than the original space, e.g., we can expect the words "Poland", "USA", "France" to have embeddings within a certain region, and the words "jump", "ride," "run" in another, distinct region;
    \item natural clustering -- categorical values could be assigned to observations within the same manifold, e.g., a region with the words "Poland", "USA", "France" can be described as  "countries";
    \item sparsity -- given an observation, only a subset of its numerical representation features should be relevant. Otherwise, we end up with complicated embeddings whose highly correlated features may lead to numerous ambiguities. 
\end{itemize}

For multimodal representation,~\cite{srivastava_multimodal_2014} identify more factors that should be taken into account: (1) similarity between individual modalities should be preserved in their joint representation; (2) robustness to the absence of some modalities -- it should be always possible to create multimodal embedding.

\subsection{Multimodal Data Fusion}

Multimodal data fusion is an approach for combining single modalities to derive multimodal representation.
A few issues should be taken into account~(\cite{gao_survey_2020}) when it comes to fusing several modalities:
\begin{itemize}
    \item intermodality -- the combination of several modalities, which leads to better and more robust model predictions~(\cite{baltrusaitis_multimodal_2019})
    \item cross-modality -- this aspect assumes inseparable interactions between modalities. No reasonable conclusions can be drawn from data unless all modalities are joined~(\cite{frank_vision-and-language_2021})
    %A perfect example is the image captioning with reading comprehension task~\cite{sidorov_textcaps:_2020};
    \item missing data -- for some cases, particular modalities might not be available. An ideal multimodal data fusion algorithm is robust to missing modalities and uses others to compensate for the information loss, widespread in recommender systems.%~\cite{wang_lrmm:_2018}.
\end{itemize}
Classically, the existing multimodal representation techniques are divided in two categories~(\cite{gallo_multimodal_2017}, \cite{baltrusaitis_multimodal_2019}): early (feature) and late  (decision) fusion. % -- Figure~\ref{fig:data-fusion}.  
In the early feature approach, all modalities are combined. This is usually achieved by concatenating their vector representations at an initial stage, and then one model is trained~(\cite{kiela_efficient_2018}). In the case of late fusion, several independent models concerning each modality are trained. Then their outputs are connected. The connection can be made arbitrarily. One can average the outputs, pick the most frequent one (in classification tasks) or concatenate them and build a model on top of that to obtain a final output~(\cite{kiela_efficient_2018}). %,~\cite{gallo_multimodal_2017}. 
Neither of these data fusion approaches can be described as the best one~(\cite{gallo_multimodal_2017}); both have been proven to yield promising results in various scenarios.

\subsubsection{Deep Learning Models}

The most popular multimodal fusion techniques are based on deep learning solutions. \cite{gao_survey_2020} describe such architecture ideas, along with cases of their most representative. Four prominent approaches are: Deep Belief Nets, %~\cite{srivastava_multimodal_2014}, 
Stacked Autoencoders, %~\cite{ngiam_multimodal_2011}, 
Convolution Networks, %~\cite{ma_multimodal_2015}, 
and Recurrent Networks. %~\cite{mao_deep_2015}. 
However, despite their promising results in the field of multimodal data fusion, deep learning models suffer from two main issues~(\cite{gao_survey_2020}). Firstly, deep learning models contain an enormous amount of free weights, especially parameters associated with a modality that brings little information. That results in high resource requirements -- an undesirable feature in a production scenario. Secondly, multimodal data usually comes from very dynamic environments. Therefore, there is a need for a flexible model that can easily adapt to all data changes. 

Enormous computational requirements and low flexibility suggest the possibility to explore other ideas applied to any task, despite its type of modalities. Furthermore, these ideas can combine with deep learning techniques and existing multimodal models to obtain state-of-the-art solutions, which would be applicable in every field, and robust to all data imperfections (missing modalities, data distribution changes over time in a production case, etc.). One of the possible solutions is the usage of hashing methods. The following section discusses the strengths and weaknesses of such algorithms.

\subsubsection{Hashing Ideas}
Another promising approach in multimodal data fusion is associated with hashing models. They work by identifying manifolds in origin space and then transforming data to lower-dimensional spaces while preserving observations' similarities. Such algorithms are capable of constructing multimodal representation on-the-fly and have been proven effective in information retrieval  problems~(\cite{cao_review_2020}), recommendation systems~(\cite{dabrowski_efficient_2020}), and object detection cases. ~(\cite{ghayoumi_local_2018}). 
The main advantages~(\cite{cao_review_2020}) of hashing methods are that they: (1) are cost-effective in terms of memory usage, (2)  detect and work within manifolds, (3) preserve semantic similarities between points, (4) are usually data-independent, (5)  are suitable for production cases as they are robust to any data changes.

Unfortunately, hashing methods struggle with one issue. The mapping of high dimensional data to much simpler representations can result in the loss of certain information about specific observations~(\cite{cao_review_2020}). Therefore, it has to be verified if hashing ideas can be applied to other fields, apart from similarity search tasks. Perhaps their ability to combine multiple modalities while maintaining low costs and robustness to data changes recompenses the lost information.

\subsubsection{Sketch Representation}
The sketch representation has already been proven effective if fed with visual, behavioral, and textual data for the recommendation and similarity search tasks~(\cite{dabrowski_efficient_2020}). 
The idea of this representation comes from combining two algorithms: Locality Sensitive Hashing and Count-Min Sketch. All modalities are transformed into mutual space with the use of hash functions. Generally, a sketch is a one-hot sparse matrix containing all combined modalities. Hash functions make the representation modality-independent, robust to missing modalities, and easily interpreted. Furthermore, modalities can be added to the sketch in an on-the-fly manner, which is extremely important in a production scenario.

In our research, we slightly modify this sketch representation to the binarized form. Instead of representing an observation with the subspace id, it can be represented as a set of binary features. Then $0$ and $1$ represent where the point lies concerning a single hyperplane. Such a sketch should preserve more information about a single observation.
\begin{figure}
    \centering
    \includegraphics[width=0.6\linewidth]{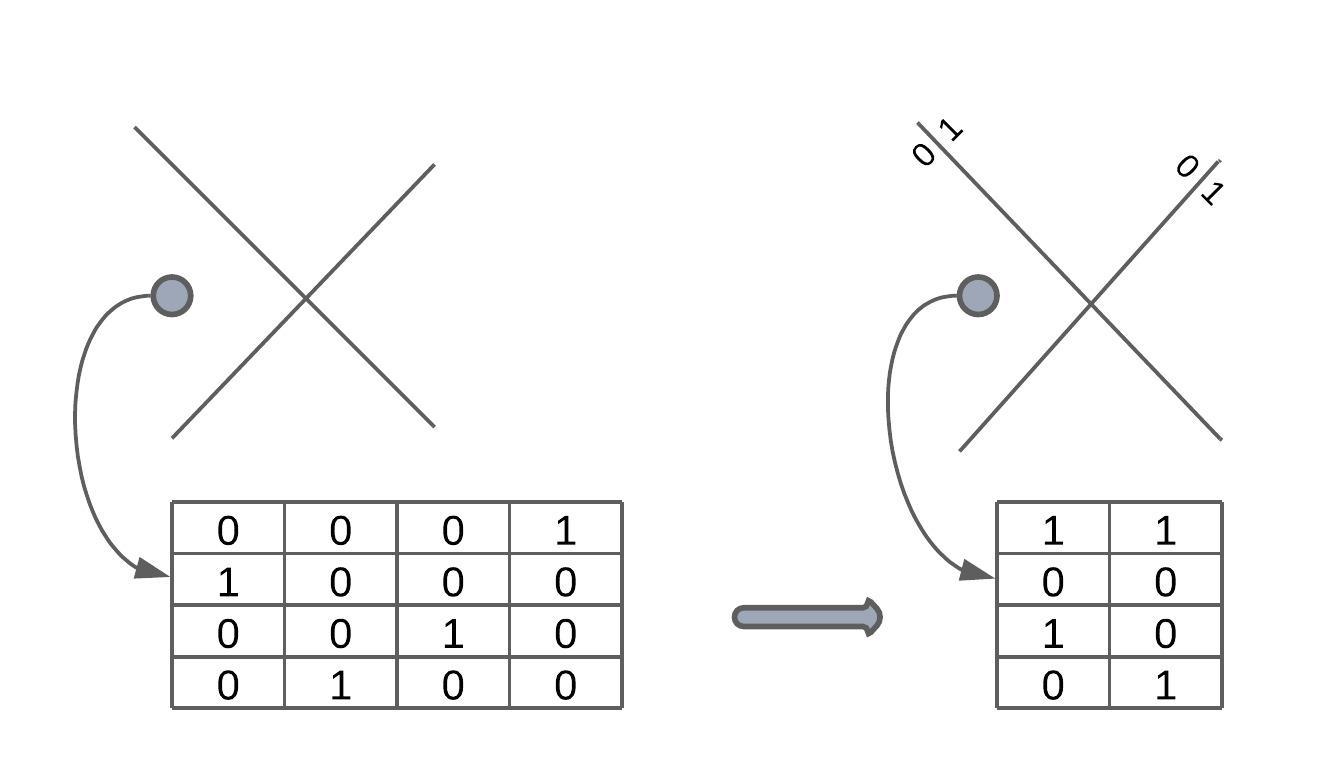}
    \caption[Sketch binarization]%
    {The idea of binarizing the sketch. Instead of representing an observation with the subspace id, it can be represented as a set of binary features. Then $0$ and $1$ represent where the point lies concerning a single hyperplane. Such a sketch consumes much less memory and perhaps preserves more information about a single observation.}
    \label{fig:binarize-sketch}
\end{figure}

\subsection{Multimodal Model Evaluation}\label{section:multimodal-evaluation}
The problem of evaluating the multimodal data fusion algorithm is not straightforward, and there does not exist any universal metric that would measure the aspect of captured inter and cross-modalities~(\cite{frank_vision-and-language_2021}). However, we can assess whether learning from multiple data types at once enhances task performance.

The most popular way of verifying the quality of the multimodal fusion model~(\cite{gallo_multimodal_2017}) is to compare its performance scores (precision, AUC, etc.) to ones achieved by models considering single modalities.  With such an approach, we can state whether and to what extent combining modalities brings new information. %Furthermore, the concept of learning multimodal data fusion is similar to the Nearest  Neighbor (NN)~\cite{roussopoulos_nearest_1995} problem. 
We also aim to preserve all similarities between observations, i.e., similar observations should be comparable in their multimodal representations. Therefore several works~(\cite{kiela_efficient_2018},~\cite{dabrowski_efficient_2020}), compare their multimodal models to NN algorithms, which serve as a good baseline.
Lastly, multimodal models should be tested on the time they need to adjust additional modalities. In certain cases~(\cite{kiela_efficient_2018}), adding new modalities slightly improves the results, while the training time increases dramatically. As a result, the model, despite its excellent performance, might be unfeasible in the production scenario. Therefore not only should we focus on the scores the model achieves, but also take into account its flexibility and simplicity.

In our work, we analyze three techniques for building the multimodal representation -- the late fusion, the early fusion, and the sketch. The main ideas of all of them are described in this section. The next part introduces datasets that are used for analyses.

\section{Datasets}\label{section:datasets}

In order to evaluate different multimodal representations, we compare them in classification problems, so we can easily assess the amount of information that each modality brings. Furthermore, we analyze scenarios, where the multimodality is expected to benefit the model, i.e. model based on several modalities should obtain better performance than the unimodal model.

Table~\ref{table:datasets-summary} presents the overall characteristics of three used datasets (Amazon Reviews, MovieLens25M, and MovieLens1M) and the employed machine learning tasks. In our experiments, Amazon Reviews and Movielens25M are split at $0.6/0.2/0.2$, the training, validation, and test set. With such division, the results on test sets are representative (12,000 samples), while maintaining enough observations for training. The split of $0.8/0.2$ -- the training and test set --  is used in MovieLens1M. Instead of one predefined validation set, MovieLens1M uses the k-fold Cross-Validation (on the training set). We opt for Cross-Validation cause this dataset is much smaller than the other two.

\begin{table}[!ht]
% \begin{minipage}{\columnwidth}
\centering
\begin{tabular}{llll}
\hline
\begin{tabular}[l]{@{}l@{}}\textbf{Dataset}\\\textbf{name}\end{tabular} & \textbf{Size} & \multicolumn{1}{l}{\textbf{Modalities}}                             & \textbf{Task}                          \\ \toprule
\begin{tabular}[l]{@{}l@{}}Amazon \\ Reviews\end{tabular} 
        & 60,000        & \begin{tabular}[c]{@{}l@{}}- textual (product description \\ \ \ and title)\\ - visual (product image)\end{tabular}                 & \begin{tabular}[l]{@{}l@{}}product classification\\ (multiclass)\end{tabular} \\ \midrule
\begin{tabular}[l]{@{}l@{}}MovieLens \\ 25M\end{tabular}  & 60,763        & \begin{tabular}[l]{@{}l@{}}- textual (movie plot)\\ - visual (movie poster)\\ - graph (movies seen by one user)\end{tabular} & \begin{tabular}[l]{@{}l@{}}genres classification\\ (multilabel)\end{tabular}  \\ \midrule
\begin{tabular}[l]{@{}l@{}}MovieLens \\ 1M\end{tabular}         & 6,040         & \begin{tabular}[l]{@{}l@{}}- textual (movie plot)\\ - visual (movie poster)\end{tabular}                                     & \begin{tabular}[l]{@{}l@{}}gender classification\\ (binary)\end{tabular}      \\ \bottomrule
\end{tabular}
% \end{minipage}
\caption[Dataset characteristics]{Dataset characteristics}%
\label{table:datasets-summary}
\end{table}

\textbf{Amazon Reviews} dataset consists of products from Amazon. The data used in experiments are only a small subset of the original dataset\footnote{jmcauley.ucsd.edu/data/amazon/} offers -- "Clothing, Shoes, and Jewelry" category. Intuitively, products within this category should be divergent enough. Furthermore, products within the fashion field can be expected to contain meaningful descriptions, titles, and, most importantly, images. 
The dataset extracted from Amazon Reviews is characterized by missing modalities. Around $25\%$ of descriptions are lost, and $3\%$ of titles are empty; however, all the records contain images. 

\textbf{MovieLens25M} dataset is the largest of the MovieLens family.\footnote{https://grouplens.org/datasets/movielens/} It consists of 25 million ratings applied to $62\,000$ movies by $162\,000$ users. Each movie is associated with genres and with one IMDb identifier. The metadata are collected based on such an identifier -- movie's poster and plot, the visual and textual embedding, respectively. Some movies are subscribed to incorrect IMDb identifiers and therefore removed. After preprocessing, the dataset in our experiment consists of $60\,763$ unique movies. The existing modalities are: movie poster (image), movie plot (text), also viewed (graph).

There are some missing modalities: every movie has a plot, $63$ movies lack the poster, and $3\,231$ movies are not rated by users in MovieLens25M. Missing modalities are proportional to class sizes, meaning the most considerable number of missing modalities is in the Drama class.
Here, classes are highly imbalanced. Around 40\% of all observations are dramas, 27\% are comedies, whereas film-noir and IMAX movies are associated with less than 1\% of observations. There is also a high number of movies that are not subscribed to any specific category -- the \textit{(no genres listed)} class. 

\textbf{MovieLens1M} is a smaller version of the previously described dataset and consists of $6\,040$ users that have rated approximately $4\,000$ movies. However, apart from users' ratings, their demographic data are available. It allows for the investigation of dependencies between watched content and some user characteristics such as gender, age, or occupation. After preprocessing for our experiment, only two modalities are utilized: textual (the plot) and visual (the poster). Again the data are downloaded based on IMDb identifiers associated with movies. There are no missing modalities. Classes are imbalanced -- around $71.7\%$ of users are males. This imbalance is taken into account during the evaluation, by using the MCC metric. 

\section{Research Design and Experiments}
\label{section:research_design}
We carry out three experiments with three classification tasks: multiclass, multilabel, binary (see Table~\ref{table:datasets-summary}). The type of task is determined by the characteristics of the dataset.

\subsection{Multiclass Classification -- Amazon Reviews Dataset}\label{section:amazon-reviews-experiments}
\begin{figure}[!ht]
     \centering
     \includegraphics[width=1\columnwidth]{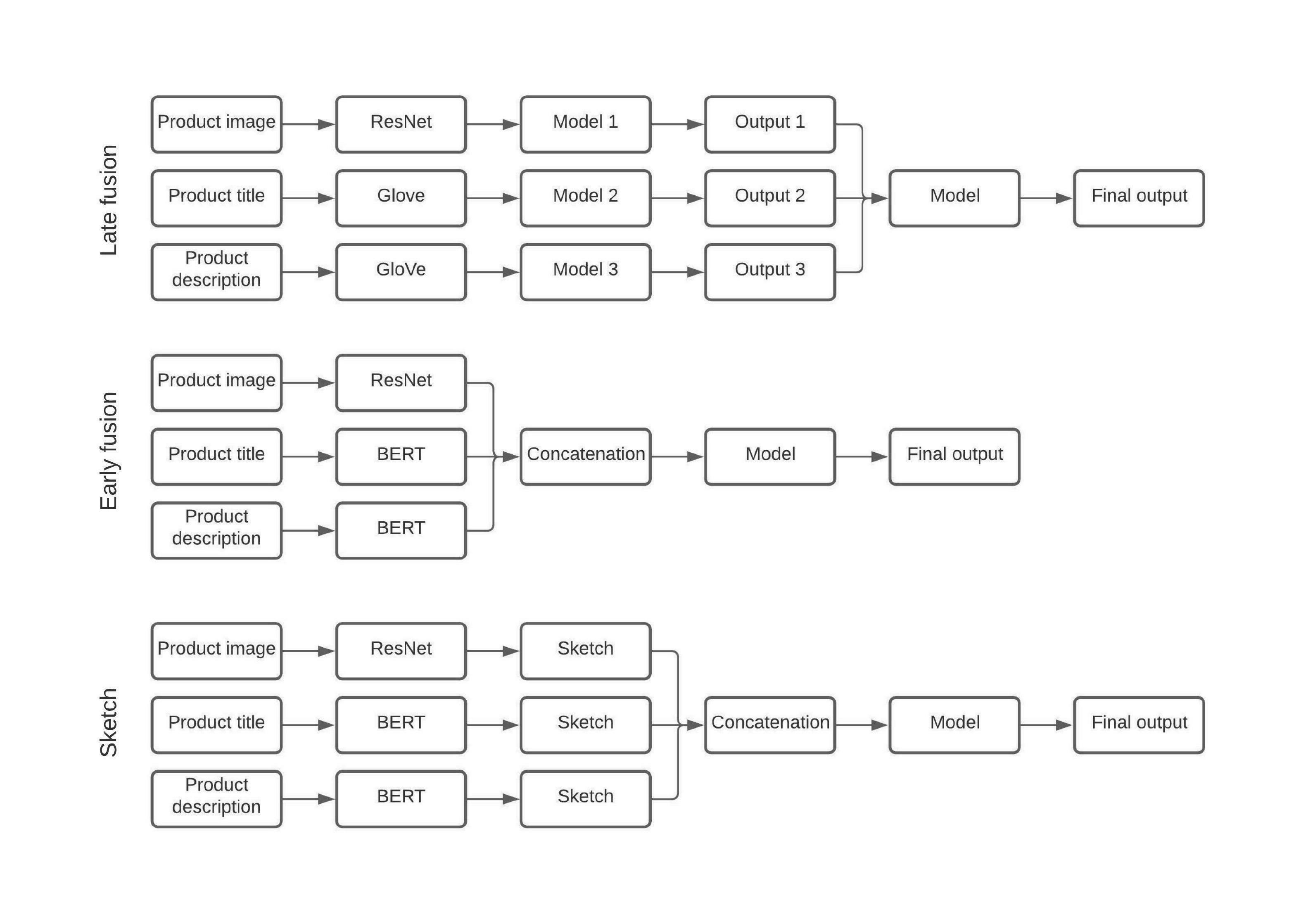}
     \caption[Amazon Reviews architectures]%
     {Here we present the architectures that are used in the Amazon Reviews Dataset. They are depicted in their trimodal forms. Bimodal and unimodal architectures look the same. The only difference is that unimodal models do not have a model on top of their outputs in the late fusion approach. Above the output means the probabilities of observations belonging to respective classes.}
     \label{fig:amazon-architecture}
 \end{figure}

Analyses on the Amazon Reviews Dataset consider the multiclass classification task based on two modalities -- textual and visual. Classes are evenly distributed, and therefore the accuracy metric is used. The primary purpose of experiments on this dataset is to compare multimodal fusion techniques based on neural networks:
 \begin{itemize}
     \item the late fusion, where modalities are treated independently. Here models' architectures are based on~\cite{wirojwatanakul_multi-label_2019}.
    \item early fusion models, where embeddings of each modality are concatenated at the input level.
    \item the sketch, including the classical and binarized version.
\end{itemize}
Their overall architectures are visible in Figure~\ref{fig:amazon-architecture}. All of these approaches use Adam as the optimizer and categorical cross-entropy loss function. Everywhere the batch size is set to $32$. All models are trained for 10 epochs, except for the binarized sketch models which are trained for 20 epochs. The learning rate differs per approach and equals $10^{-4}, 10^{-3}, 10^{-5}, 10^{-4}$ in the late fusion, the early fusion, the classical, and the binarized sketch approach, respectively.

\subsubsection{Late Fusion Models}
While deploying ideas from~\cite{wirojwatanakul_multi-label_2019}, slight modifications are required to transfer from multilabel to multiclass classification task: a different loss function or the number of used neurons within layers. Both description and title are represented with GloVe. %~\cite{pennington_glove:_2014}. 
Image embeddings are created with ResNet50. All multimodal models are identical. The outputs from unimodal models are fed into a dense layer, with 20 units and ReLU as the activation function. Finally, a 12-unit dense layer for product classification is added. Therefore, the only difference between bimodal and trimodal models is the input size, equal to 24 and 32.

\subsubsection{Early Fusion} Here each modality is firstly transformed into a numerical vector with BERT (\textit{bert-base-uncased}) for both titles and descriptions, resulting in $768$-dimensional vectors. 
For image embeddings, ResNet50 is used. The last layer is removed, which results in $2048$-dimensional vectors.
In this approach, models have the same architecture, whether they are uni or multimodal. Therefore the only difference is that in the multimodal case, appropriate embeddings are  concatenated, resulting in $1\,536$ (textual + textual), $2\,816$ (textual + image), $3\,584$ (trimodal) input embedding sizes.
The model architecture consists of three blocks: dense-dropout layers, with 64 units, ReLU as the activation function, and the dropout layer -- rate is set to $0.1$. Subsequently, one dense layer is added, also with 64 units and ReLU. However, in this case, the dropout is not applied. Finally, there is the 12-unit dense layer with softmax, responsible for classification.

\subsubsection{Sketch Representation} Similarly to the early fusion approach, BERT is used for the textual and ResNet50 for the visual data. Firstly, embeddings are transformed into sketches. The sketch's depth equals $128$, and the width equals $512$. Then sketches are flattened and concatenated in multimodal cases. 
The architecture is universal, despite the number of modalities. The model starts with two identical blocks consisting of a dense layer, a dropout layer, and batch normalization. The dense layer in the first block consists of $1\,024$ neurons and the second of $512$ neurons. After the second block, there is a $128$-dimensional dense layer followed by a dropout layer. All of these dense layers use ReLU, and the dropout rate is set to $0.2$. Finally, there is a $12$-dimensional dense layer with softmax, responsible for classification. To provide a fair comparison with the classical sketch, the binarized version uses the same architecture.

\subsection{Multilabel Classification -- MovieLens25M}\label{section:movielens25m-experiments}
Movielens25M dataset considers multilabel genre classification; its classes are highly imbalanced and correlated. Therefore, threshold-independent metrics are used for evaluation: Area Under Curve (AUC) and mean Average Precision (mAP). Such an approach makes analyzed solutions independent from class threshold selection at testing time. Furthermore, to take the class imbalance into account, \textit{micro} versions of these metrics are used: micro-AUC and micro-mAP.

\begin{figure}[!ht]
    \centering
    \includegraphics[width=1\columnwidth]{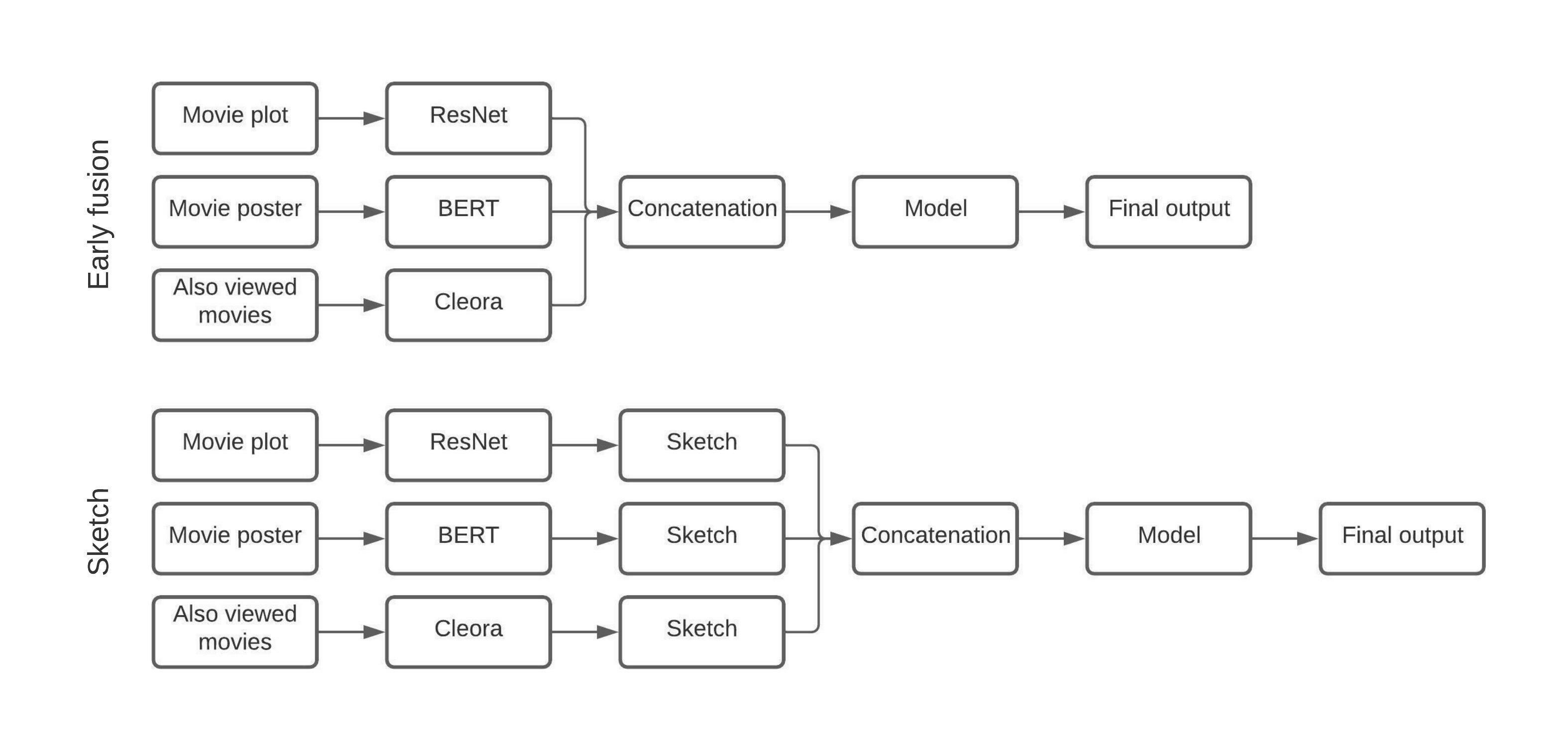}
    \caption[MovieLens25M architectures]%
    {Architectures used in MovieLens25M dataset.}
    \label{fig:movielens25m-architecture}
\end{figure}

Here only the sketch and the early fusion representations are considered. Figure~\ref{fig:movielens25m-architecture} illustrates both of the approaches.
Movielens25M encapsulates three kinds of modalities: graph, text, and images. The first one is encoded with Cleora~(\cite{rychalska_cleora:_2021}). Cleora's hyperparameters are set as follows: the embedding size - $1\,024$, the number of iterations - 1. Such parameters are optimal in terms of preserving strong similarities between entities. Text data are represented with BERT (\textit{bert-base-cased}) and visual data with ResNet50. Both sketch's hyperparameters (depth, width) are set to $128$.

Models consist of four dense layers. The first three use ReLU and have $1\,024$, $512$, $128$ neurons, respectively. The last 20-dimensional layer is used for multilabel classification and therefore uses sigmoid. Models use Binary Cross-Entropy loss function, the Adam optimizer with a learning rate equal to $1^{-5}$, and its other parameters are set as default. Each model is trained for 20 epochs, apart from multimodal early fusion models and unimodal graph models that are trained for $30$ epochs. The batch size is set to 64. Every model is trained $5$ times using train and validation data and evaluated on the test data to obtain robust results. 

\subsection{Binary Classification -- MovieLens1M}
\label{section:movielens1m-experiments}
The task on MovieLens1M is a binary classification -- predicting people's gender based on their behavioral data (here, movies they watched). 

\begin{figure}[!ht]
    \centering
    \includegraphics[width=1\linewidth]{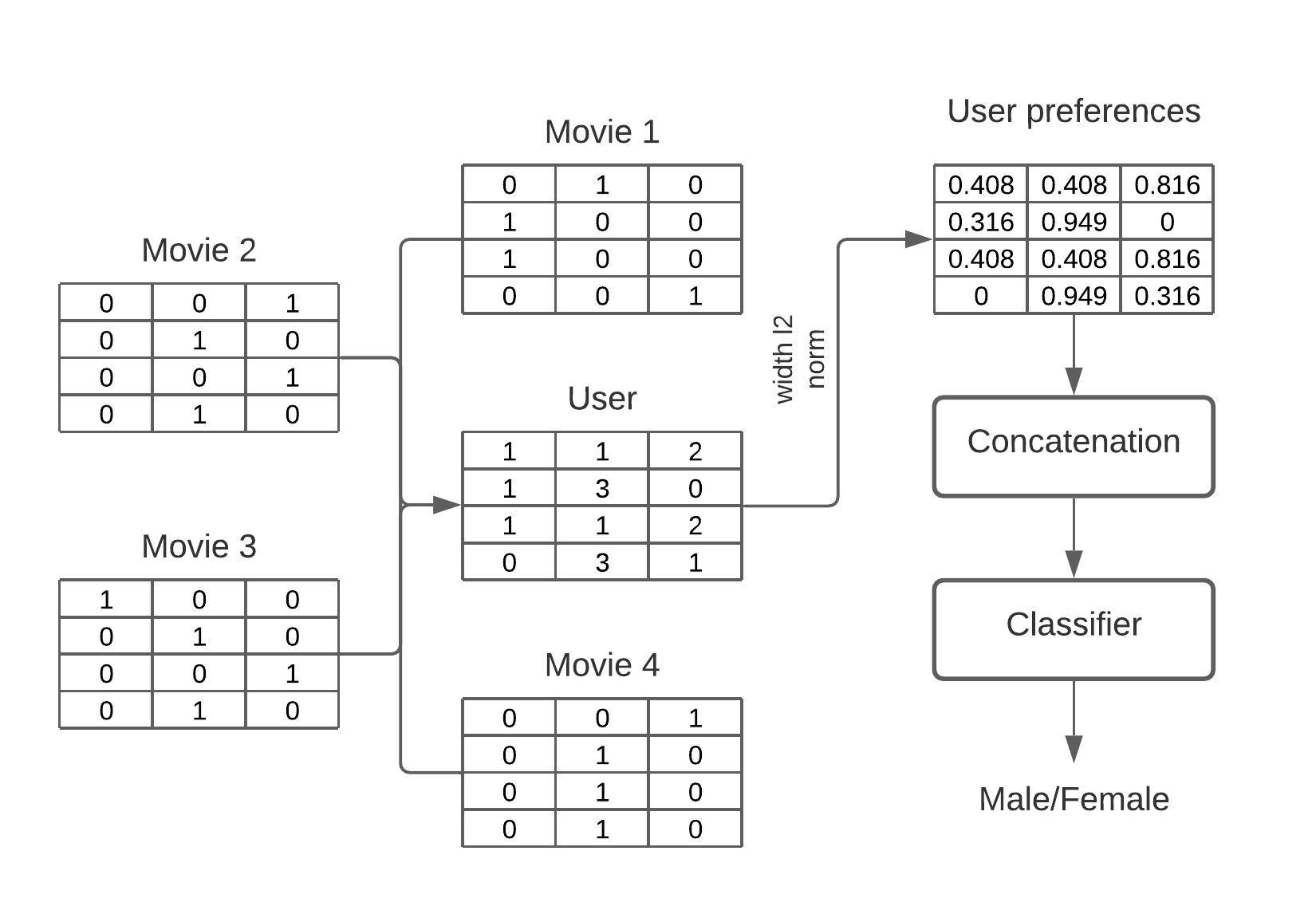}
    \caption[Gender classification architecture]%
    {The architecture of the gender classification. The user is represented as a sum of sketches. Then the width-wise L2 normalization is performed. Finally the sketch is flattened and fed into the classifier.}
    \label{fig:movielens1m-architecture}
\end{figure}

Here, in the sketch representation, every movie is represented with the sketch and the user as a sum of such sketches, see Figure~\ref{fig:movielens1m-architecture}. However, unlike the recommender scenario, the output is a number between 0 and 1, representing the probability that the user is a male or a female.

Every user has rated (watched) at least 20 movies. Movies are represented with the plot and the poster, the same way as in MovieLens25M. Then, users are represented with the content they watch. Once a user sketch is created, it is flattened and fed into Logistic Regression with regularization parameter $C$ set to $0.1$. The sketch's width equals 512, while the depth is adjusted to the modality type -- it equals 210 for the movie plot, and 128 for the movie poster. In the bimodal model, the flattened sketches are concatenated. Because of the class imbalance, the Matthew Correlation Coefficient (MCC) is used for the evaluation. Movie plots are embedded with BERT (\textit{bert-base-cased}), and posters are embedded with ResNet50.

\section{Results}
\label{section:results}
Tables~\ref{table:amazon-reviews-results},~\ref{table:movielens25m-results}, and~\ref{table:movielens1m-results} shows our results for different tasks and datasets. We can see that only in one dataset -- Amazon Reviews -- do all the used modalities improve the final results. Other tests on MovieLens datasets show consequently that posters reduce the results, and intuitively we can imagine that the posters do not have much substantial information about movies. %In the case of the Amazon Reviews dataset, the multimodal model outperformed the best unimodal model by $0.05$. In contrast, the best multimodal model in the MovieLens25M dataset exceeded the unimodal models' results by $0.011$ and $0.043$ in the AUC and mAP metrics, respectively.

\begin{table}[!ht]
\footnotesize
\begin{center}
\begin{tabular}{@{\extracolsep{\fill}}ccccc@{\extracolsep{\fill}}}
\toprule% Modality
Modality & Late fusion & Early fusion & Sketch & Sketch binarized \\ \midrule%
\textbf{title} & { 0.919$\ \pm\  $0.002}  & { 0.857$\ \pm\  $0.002}  & { 0.780$\ \pm\  $0.002} & {0.822$\ \pm\  $0.004} \\ 
\textbf{description} & { 0.706$\ \pm\  $0.003} & { 0.662$\ \pm\  $0.001} & { 0.613$\ \pm\  $0.003} & {0.636$\ \pm\  $0.004}\\
\textbf{image} & { 0.836$\ \pm\  $0.005} & { 0.851$\ \pm\  $0.002} & { 0.770$\ \pm\  $0.002} & {0.812$\ \pm\  $0.003}\\ 
 \textbf{title \& description} & { 0.962$\ \pm\  $0.001} & { 0.910$\ \pm\  $0.002} & { 0.851$\ \pm\  $0.002} & {0.868$\ \pm\  $0.006}\\
\textbf{image \& description} & { 0.927$\ \pm\  $0.001} & { 0.898$\ \pm\  $0.003} & { 0.847$\ \pm\  $0.002}
& {0.859$\ \pm\  $0.003}\\
\textbf{image \& title} & { 0.962$\ \pm\  $0.001} & { 0.930 $\ \pm\  $0.002} & { 0.885$\ \pm\  $0.002}
& {0.903$\ \pm\  $0.004}\\
\textbf{image \& title \& description} & \textbf{ 0.969$\ \pm\  $0.001} & \textbf{ 0.940$\ \pm\  $0.002} & \textbf{ 0.912$\ \pm\  $0.002} & \textbf{0.914$\ \pm\  $0.002}\\

\bottomrule
\end{tabular}
\caption[Amazon Reviews Dataset -- results]{Amazon Reviews Dataset -- the accuracy of each approach, for each modality combinations. Results are averaged over 10 runs.}\label{table:amazon-reviews-results}
\end{center}
\end{table}

\begin{table}[!ht]
\centering
\footnotesize
\begin{tabular}{@{\extracolsep{\fill}}ccccc@{\extracolsep{\fill}}}
\toprule% Modality
& \multicolumn{2}{c}{\textbf{Early fusion approach}} & \multicolumn{2}{c}{\textbf{Sketch approach}} \\
\multicolumn{1}{c}{\textbf{Modality}} & \textbf{AUC} & \textbf{mAP} & \textbf{AUC} & \textbf{mAP} \\ \midrule%

\textbf{text} & 0.907 $\pm$ 0.001 & 0.549 $\pm$ 0.001 & 0.877 $\pm$ 0.003 & 0.465 $\pm$ 0.003 \\ 
\textbf{image} & 0.815 $\pm$ 0.002 & 0.350 $\pm$ 0.001 & 0.778 $\pm$ 0.003 & 0.297 $\pm$ 0.003 \\
\textbf{graph} & 0.824 $\pm$ 0.001 & 0.382 $\pm$ 0.002 & 0.797 $\pm$ 0.003 & 0.323 $\pm$ 0.003 \\
\textbf{text \& graph} & \textbf{0.918 $\pm$ 0.001} & \textbf{0.592 $\pm$ 0.002} & \textbf{0.883 $\pm$ 0.002} & \textbf{0.484 $\pm$ 0.002} \\
\textbf{text \& graph \& images} & 0.915 $\pm$ 0.001 & 0.583 $\pm$ 0.002 & 0.880 $\pm$ 0.002 & 0.482 $\pm$ 0.003 \\ 
\bottomrule
\end{tabular}
\caption[MovieLens25M -- results]{Results of the early fusion and the sketch on MovieLens25M. Results are averaged over 5 runs.}\label{table:movielens25m-results}
\end{table}

\begin{figure}[!ht]
    \centering
    \includegraphics[width=\textwidth]{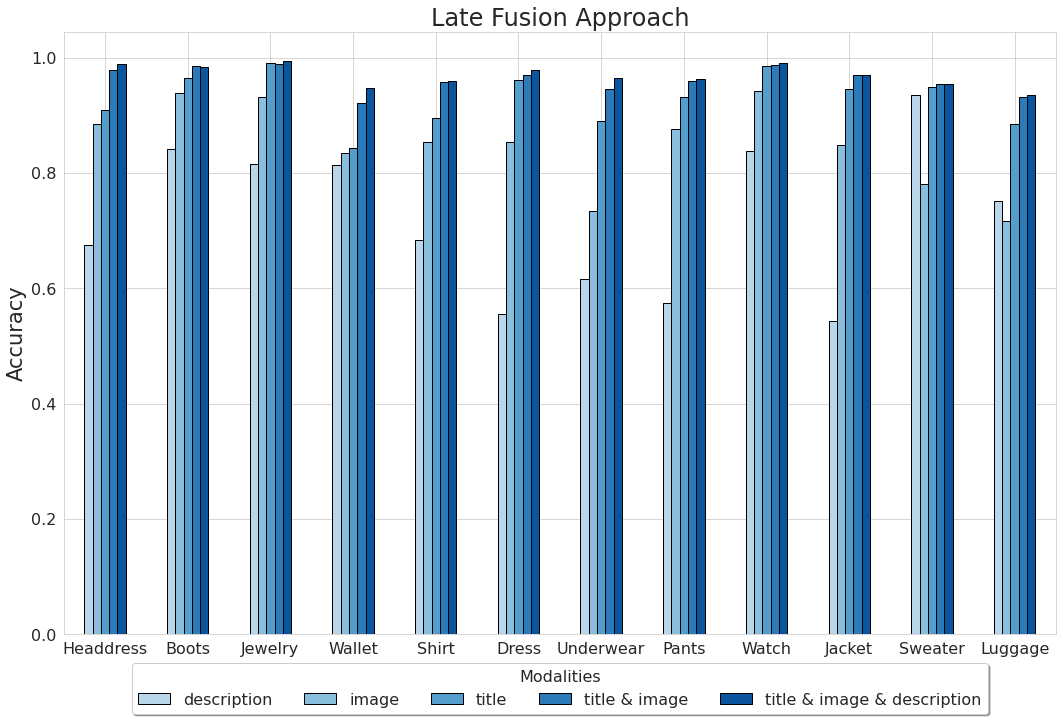}\hfill
    \caption[Amazon Reviews -- late fusion results.]%
    {The late fusion models results with respect to categories.}
    \label{fig:amazon_reviews_article_classes}
\end{figure}

Since the impact that each modality has on the model's performance is not so clear on Amazon Reviews Datasets, we decided to explore them thoroughly with respect to each category, see~Figure \ref{fig:amazon_reviews_article_classes}. The analyzed results come from the late fusion approach, which achieved the best results.

The bimodal and trimodal models outperform unimodal models in every category (accuracy metric). The most significant benefit from exploiting multiple modalities can be noticed within classes, where unimodal models yield close results, for instance, the \textit{Wallet} class. On the other hand, categories, where one of the unimodal approaches is undeniably better, do not gain much from multimodality -- the \textit{Watch} class.

Surprisingly, despite the absence of around 25\% of the data concerning the \textit{Sweater} category, the description model significantly outperforms the image model within this class. This example shows that each modality, even the one with missing data, can yield unique information and be helpful until proven otherwise.

In most cases, the trimodal model does not significantly outperform the bimodal, based on title and image. Furthermore, training the trimodal model takes twice the time the bimodal model needs. 
% In the case of this experiment, it is not relevant since it concerns training models on $48\,000$ observations.
In the real-world scenario, the trimodal model could be too expensive for the slight improvement it provides. Then, only titles and images could be considered valuable.

In the case of gender classification, the highest MCC -- $0.543$ -- indicates that the proposed architecture might be successfully used to solve demographic classification problems. It could be a powerful tool, for instance, for e-commerce or user-generated platforms. By modeling users' characteristics, companies cold personalize their offers to maximize their profits. On the other hand, users will benefit from personalized content. However, more experiments on larger datasets should be performed to confirm the usefulness in the production scenario.

\begin{table}[!ht]
\footnotesize
\begin{minipage}{\columnwidth}
\centering
\begin{tabular}{cc}
\toprule
\textbf{Modality} & \textbf{Matthew Corellation}\\
\midrule
\textbf{Movie poster} & 0.518 \\
\textbf{Movie plot} & \textbf{0.543} \\
\textbf{Movie plot \& poster} & 0.532 \\
\bottomrule
\end{tabular}
\caption[MovieLens1M -- gender classification results]{Results of the gender classification on test set -- Logistic Regression.}\label{table:movielens1m-results}%
\end{minipage}
\end{table}

% TA CZĘŚĆ JEST DODANA
To the best of the authors' knowledge, our work, for the first time, classifies users based on the content of their interests (see experiment on MovieLens1M for gender classification). So far, the existing approaches~(\cite{de_cnudde_benchmarking_2020}) %,~\cite{liu_efficient_2019} 
have considered only the behavioral data, which in the case of MovieLens1M are present as ratings that users gave to movies. Our experiment shows that it is possible to model user characteristics with the content in which the users are interested. Though only simple classifiers were tested, it was proven that our approach is reasonable. 

All the tests show that combining several modalities boosts model performance; however, the data must be meaningful for the given task. If possible, consideration of additional modalities is suggested, as they may provide new insight into the analyzed data. Our research concludes that multiple data modalities should be utilized cautiously, as some might yield little information (here, movie posters are redundant for genres and gender classification). 

% TA CZĘŚĆ JEST DODANA

% podkreślić wskazówkę jaką może byc tablea/artykuł dla przyszłych badań 
The main conclusions that can be drawn from tests are summarized in Table~\ref{table:techniques-suggestions}. These conclusions are made upon the following criteria: the influence that each modality has on the analyzed machine learning (ML) problem; the type of the ML task; the memory constraints while training and predicting phase.

\begin{table}[!ht]
\centering
\footnotesize
% \begin{minipage}{\textwidth}
\begin{tabularx}{\textwidth}{lX}
\toprule
\textbf{Technique} & \textbf{Notes and recommendations:} \\
\toprule
\multirow{3}{*}{\textbf{Late fusion}} &
\tabitem use when one modality is dominant (its unimodal model gives significantly better results than others) \\
& \tabitem use when every unimodal model achieves high performance \\
& \tabitem preferred in classification tasks, in which probabilities from unimodal models can be easily combined, such as in~\cite{wirojwatanakul_multi-label_2019}  \\
\midrule
\multirow{3}{*}{\textbf{Early fusion}}  &
\tabitem use when modalities are dependent, i.e. no reasonable conclusions can be made unless modalities are studied together (e.g. questions to pictures~\cite{singh_towards_2019}) \\
& \tabitem use when all unimodal models yield similar results \\
& \tabitem easy to use pre-trained models; the model input is their early fusion \\
\midrule
\multirow{3}{*}{\textbf{Sketch}} &
 \tabitem memory efficient (sketches can be very short vectors, can be stored  and used efficiently, with low memory usage) \\
 & \tabitem suggested for information filtering problems, such as recommender systems~(\cite{dabrowski_efficient_2020}) \\
 & \tabitem easy to use pre-trained models as inputs to sketching technique \\
\bottomrule
\end{tabularx}
\caption[Comparing techniques selection criteria for building multimodal representation]{The recommendations are constructed with respect to three criteria:  the influence that each modality has on the analysed machine learning (ML) problem; the type of the ML task; the memory  constraints while training and predicting phase}\label{table:techniques-suggestions}%
% \end{minipage}
\end{table}

Tests conducted on Amazon Reviews prove that the best one is the late fusion in nearly all scenarios. Here, the reason is probably the type of textual data. Many descriptions and titles contain words that are sufficient on their own for a good prediction. Therefore in situations where one of the modalities is dominant, we believe that the best approach would be the late fusion. The model efficiently exploits the information from the most informative modality and can expand it using additional modalities.   

On the other hand, the early fusion approach can reveal information from the combination of the modalities (interaction between modalities), which can be interdependent. Therefore in scenarios, where several modalities affect the model equally, their interactions might reveal hidden information, and hence the early fusion should be chosen.

Our experiments show that the sketch performs worse in typical classification problems. However, this representation can be the least consuming in storing the data representations. Furthermore, when the number of used modalities increases, the performance of sketch-based model almost always increases. Possibly, this improvement could be maintained until reaching the state-of-the-art performance. If such behavior is confirmed, the sketch representation could offer ground-breaking perspectives in the world of Big Data -- highly accurate models with optimized storage consumption.

\section{Conclusions}
\label{section:conclusions}

% In this research we explored three techniques for multimodal fusion, the late fusion, the concatenation, and the sketch.
% We outlined a few aspects (see table~\ref{table:techniques-suggestions}) which can be helpful in future works on multimodal problems.

% Furthermore, we identified three main factors that should be taken into account while building multimodal representation: the impact each modality has on the task; the type of the task; the amount of used memory, the speed of training and prediction.
In this research, we compared three techniques for multimodal fusion, the late fusion, the early fusion, and the sketch, and identified their strengths and weaknesses. 

This work does not entirely exploit the problem of choosing the proper technique for building the multimodal representation. Having such a technique, one could successfully utilize the information that comes with all available modalities, hence, obtaining the best possible performance in the considered task. 

However, we confirmed that such a choice matters, and outlined the issues that come with it.  Namely,  we distinguished three main selection criteria for choosing a technique appropriate to the ML task (see Table~\ref{table:techniques-suggestions}) that should be taken into account while building multimodal representation: the impact each modality has on the task; the type of the task; the amount of used memory. Furthermore, using these criteria we summarized the three explored techniques, the late fusion, the early fusion, and the sketch, deriving recommendations that can be used by all scientists that will be working on any multimodal task. 

We believe that the next step for comparative analysis is to create a benchmark for multimodal datasets, that would confirm all conclusions made so far in this, previous, and future works on the problem of multimodal representation. 

\section*{Acknowledgements}
The research was funded by the Centre for Priority Research Area Artificial Intelligence and Robotics of Warsaw University of Technology within the Excellence Initiative: Research University (IDUB) programme (grant no 1820/27/Z01/POB2/2021).

\bibliographystyle{unsrtnat}
% \bibliography{references}  

\end{document}